# Impact of Employing Weather Forecast Data as Input to the Estimation of Evapotranspiration by Deep Neural Network Models[**]

Pedro J. Vaz[1,4 \[0000-0002-8819-3243\]], Gabriela Schütz[1,2\[0000-0001-5081-3913\]], Carlos Guerrero[1,3\[0000-0001-9907-8235\]], and Pedro J. S. Cardoso[1,4\[0000-0003-4803-7964\]]

[1] Universidade do Algarve, Portugal
[2] CEOT – Centre for Electronics, Optoelectronics and Telecommunications
[3] MED – Mediterranean Institute for Agriculture, Environment and Development
[4] LARSyS – Laboratory of Robotics and Engineering Systems
{pjmartins,gschutz,cguerre,pcardoso}@ualg.pt

**Abstract.** Reference Evapotranspiration ($ET_0$) is a key parameter for designing smart irrigation scheduling, since it is related by a coefficient to the water needs of a crop. The United Nations Food and Agriculture Organization, proposed a standard method for $ET_0$ computation (FAO56PM), based on the parameterization of the Penman-Monteith equation, that is widely adopted in the literature. To compute $ET_0$ using the FAO56-PM method, four main weather parameters are needed: temperature, humidity, wind, and solar radiation (SR). One way to make daily $ET_0$ estimations for future days is to use freely available weather forecast services (WFSs), where many meteorological parameters are estimated up to the next 15 days. A problem with this method is that currently, SR is not provided as a free forecast parameter on most of those online services or, normally, such forecasts present a financial cost penalty. For this reason, several $ET_0$ estimation models using machine and deep learning were developed and presented in the literature, that use as input features a reduced set of carefully selected weather parameters, that are compatible with common freely available WFSs. However, most studies on this topic have only evaluated model performance using data from weather stations (WSs), without considering the effect of using weather forecast data. In this study, the performance of authors' previous models is evaluated when using weather forecast data from two online WFSs, in the following scenarios: (i) direct $ET_0$ estimation by an Artificial Neural Network (ANN) model, and (ii) estimate SR by (another) ANN model, and then use that estimation for $ET_0$ computation, using the FAO56-PM method. Employing data collected from two WFSs and a WS located in Vale do Lobo, Portugal, the latter approach achieved the best result, with a coefficient of determination ($R^2$) ranging between 0.893 and 0.667, when considering forecasts up to 15 days.

---

[**] This work was supported by Project GSSIC—Green Spaces SMART Irrigation Control under Grant ALG-01-0247-FEDER-047030.

# 1 Introduction

One strategy to create a water-efficient irrigation system is to use soil humidity sensors to maintain humidity levels between the field capacity (FC) and the management allowable depletion (MAD), where MAD is a percentage of the soil's available water-holding capacity [1]. However, implementing these systems on public green spaces can be expensive due to the cost of components, installation, and management. Vandalism and theft can also be common issues, as these locations often lack comprehensive security or surveillance. Moreover, the soil's available water-holding capacity varies based on the soil type [2], requiring the use of laboratory analysis of samples to determine the humidity values corresponding to the field capacity and wilting point (WP) for a specific space.

Crop evapotranspiration ($ET_c$), also referred to as crop water use, is the water consumed by crops. The Food and Agriculture Organization of the United Nations (FAO) recommends using the FAO-56 Penman-Monteith (FAO56-PM) formula to calculate reference evapotranspiration ($ET_0$), which is related to crop evapotranspiration through a crop coefficient ($K_c$), i.e., $ET_c = K_c ET_0$ [3]. The FAO56-PM formula considers four main meteorological parameters: temperature, humidity, wind, and solar radiation (SR). The later parameter, SR, has been identified as the primary factor influencing $ET_0$ in several computational studies [4], but its measurement requires specialized sensors such as pyranometers, which are normally associated with expensive weather stations (WSs), and require specialized maintenance and calibration [5]. Furthermore, solar radiation forecast application programming interfaces (APIs) are also not widely available or can add significant costs to a system.

This paper is part of a framework for computing optimal irrigation schedules for crops, with a focus on green spaces. The framework uses computational models, based on machine learning [6] and deep learning [7], to estimate $ET_0$ using meteorological data from on-field weather stations and meteorological data (and forecasts) from internet weather forecast services (WFSs). This aims to optimize water and energy expenditure, improve crop health, reduce reaction time in addressing problems, enhance anomaly detection methods, and maintain the quality of green spaces. The framework, called Green Spaces SMART Irrigation Control (GSSIC), is being developed as an intelligent irrigation solution that is technologically differentiated from other platforms on the market.

Vaz *et al.* [6, 7] used data from a WS in Vale do Lobo, Portugal, to explore the use of machine learning and deep learning for $ET_0$ estimation. They concluded that the best results were obtained by using a deep learning regressor to estimate solar radiation ($SR^{ANN}$), using as input a limited set of meteorological features, and then using the SR estimate along with temperature, humidity, and wind speed as input to the FAO56-PM equation ($ET_0^{Hyb}$.). Using data from the Vale do Lobo WS, this approach achieved an $R^2$ value of 0.977, a root mean square error of 0.256 mm/day, a mean square error of 0.066 mm²/day, a mean absolute error of 0.16 mm/day, and a mean absolute percentage error of 5.05 %. In addition, good results were also obtained when directly estimating $ET_0$ using an Artificial Neural Network (ANN), $ET_0^{ANN}$, where an $R^2$ value of 0.959, a root mean square error of 0.342 mm/day, a mean square error of 0.117 mm²/day, a mean absolute error of 0.25 mm/day, and a mean absolute percentage error of 7.54 % was achieved [7]. Fig. **1** shows the workflow of the two $ET_0$ estimation approaches that were just detailed.

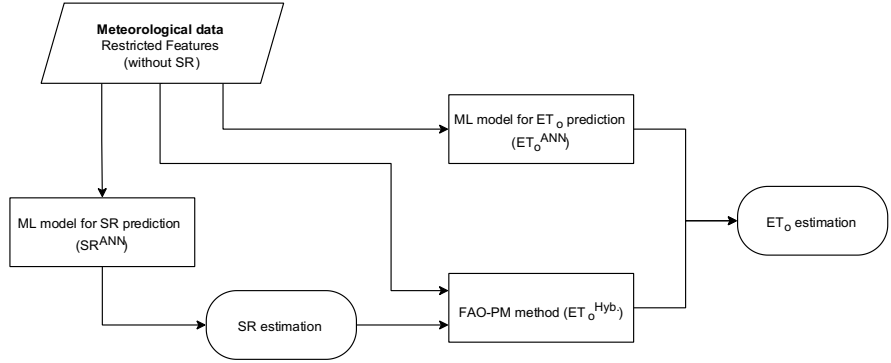

**Fig. 1.** Workflow of the two $ET_0$ estimation approaches: (i) $ET_0$ estimation model using a limited set of features ($ET_0^{ANN}$), and (ii) SR estimation model using a limited set of features ($SR^{ANN}$), and $ET_0$ computation based on using the estimated SR and a limited set of features as input to FAO56-PM formula ($ET_0^{Hyb.}$).

Importantly, the limited set of meteorological features that were used in both previously mentioned models are compatible with freely available WFSs, since SR is not required as an input feature. The current study builds on this work, by evaluating the performance of the deep learning models presented in [7], when using as input features historical weather forecast data gathered from two online WFSs, specifically Visual Crossing[1] (VC) and OpenWeatherMap[2] (OWM), for the Vale do Lobo Coordinates, allowing the assessment of the impact of using weather forecast data as input to the models.

The main contribution of this paper is the evaluation of the impact of using up to 15-day weather forecast data from online WFSs, as input to state-of-the-art $ET_0$ estimation models that are based on deep learning regressors [7]. As already described, two $ET_0$ estimation models are considered: one that directly estimates $ET_0$ from a set of restricted features ($ET_0^{ANN}$), and a hybrid model that uses a deep learning regressor to estimate SR and then applies the FAO56PM formula to compute $ET_0$ ($ET_0^{Hyb.}$). The key results are: (i) the $ET_0^{Hyb.}$ model performs better than the $ET_0^{ANN}$, (ii) both online WFSs that were considered (VC and OWM) give good and similar results for $ET_0$ estimation, and (iii) online weather forecasts can be used as inputs to the models and make $ET_0$ estimations of up to at least 11 days ($d_{11}$), at which point the $R^2$ metric falls below 0.7.

---

[1] https://www.visualcrossing.com
[2] https://openweathermap.org

This paper is structured as follows. The next section presents a brief related work review of $ET_0$ estimation using observed and forecast data. The third section explores the used datasets, which includes data from a local WS and from two WFSs (VS and OWM), ending with the experimental setup details. Section 4 explores the impact on the performance of the proposed $ET_0$ and SR regressors when using up to 15-day forecast data. Finally, some conclusions and future work are presented in the last section.

## 2 Related work

Reference evapotranspiration, $ET_0$, refers to the amount of water evaporated and transpired by a reference hypothetical grass surface with specific characteristics, including a uniform height of 0.12 m, a surface resistance of 70 sm$^{-1}$, and an albedo (reflection coefficient) of 0.23 [3]. Crop evapotranspiration, $ET_c$, represents the water needs of crops and is proportional to reference evapotranspiration through the crop coefficient, $K_c$ [3]. Therefore, accurate prediction of $ET_0$ is crucial for smart irrigation scheduling as it determines the amount of water needed to be replenished during irrigation [8]. There are several factors that set $ET_c$ apart from $ET_0$, including the crop cover density and total leaf area, the resistance of the foliage epidermis and soil surface to the flow of water vapor, the aerodynamic roughness of the crop canopy, and the reflectance of the crop and soil surface to short wave radiation [9]. With the value of $K_c$ known, the $ET_c$ can be estimated by $ET_c = K_c ET_0$.

It becomes evident that to accurately predict crop water requirements, accurate estimation of $ET_0$ is crucial. Over the years, numerous deterministic methods have been devised to estimate reference evapotranspiration by utilizing either single or limited weather parameters and are generally classified into temperature-based, radiation-based, or combination-based methods [5]. Some of the more precise ones are the methods that combine temperature and radiation, such as the Penman [10], the modified Penman [11], and the FAO-56 Penman-Monteith (FAO56-PM) [3] formulas. Shahidian et al. [5] provide a comprehensive review of several of these methods, comparing their performance under different climate conditions. The authors found, that when applied to climates differing from the ones in which the methods were developed and tested, many of these methods showed poor performance and may necessitate the adjustment of empirical coefficients to suit local climates, which is not desirable.

Recently, machine and deep learning have been used as an alternative to estimate $ET_0$. Chia, Huang, and Koo [12] found that machine learning is a promising solution for $ET_0$ estimation using common meteorological data. Granata [13] compared three evapotranspiration models and applied four machine learning algorithms (M5P Regression Tree, Bagging, Random Forest, and Support Vector Machine) to each model. The best results achieved a coefficient of determination of 0.987 and a mean absolute error of 0.14 mm/day. However, all models used net solar radiation as an input variable. Granata [14] also studied three recurrent neural network models for short-term evapotranspiration prediction, using Long Short-Term Memory (LSTM) and nonline-ar autoregressive network with exogenous inputs algorithms. Ferreira et al. [8] compared six empirical reduced set equations with ANN and Support Vector Machine (SVM) models using data from 203 weather stations in Brazil for daily $ET_0$ estimation. They found that ANN was the best-performing model when using data from up to four previous days as inputs, with a median $R^2$ value of around 0.80 for all stations.

Yang et al. [15] discuss the use of the Reduced-set Penman-Monteith model for short-term daily $ET_0$ forecasting in eight weather stations in China. The model uses temperature data from 7-day public weather forecasts and wind speed data with four different types of wind speed inputs (default value, forecasted, long-term daily average, and annual average). For example, for one of the stations the mean absolute error for 7-day $ET_0$ prediction ranged from 0.64 to 0.85 mm/day. Traore et al. [16] applied four ANN algorithms to predict $ET_0$ in Dallas, Texas, by using restricted climate information messages retrieved from a public 15-day weather forecast source. The best $ET_0$ performing model obtained a mean absolute error ranging from 0.767 to 0.996 mm/day, using as input features maximum and minimum temperature, and solar radiation. Luo et al. [17] tested four ANN models for $ET_0$ prediction using a 7-day temperature forecast from a public WFS. The obtained $R^2$ for 7-day $ET_0$ prediction, using weather forecast data as input to the models, ranged from 0.46 to 0.61.

## 3   Datasets, exploratory analyses, and experimental setup

This section introduces the datasets used in this study with particular emphasis on the exploratory data analysis over the forecast data. It ends with a summary of the experimental setup.

### 3.1   Datasets

Data was collected from a WS located in Vale do Lobo, in south Portugal, as well as historical weather forecasts from the VC and OWM WFSs, using the Vale do Lobo WS coordinates, for years 2020 and 2022. We are not considering data from 2021 as one of the services had a gap of more than three months for our location over that year. The WS is equipped with sensors from Davis Instruments, which periodically measure and record various weather parameters at a daily resolution, including: temperature (minimum, maximum, and average), relative humidity (minimum, maximum, and average), solar radiation (maximum and average), wind speed (minimum, maximum, and average), atmospheric pressure (minimum, maximum, and average), rain intensity, and precipitation. The collected historical weather forecast data (from VC and OWM WFSs) includes 15-day historical forecasts, d0, d1, ..., d15, where d0 represents today's forecast that was generated today (at midnight), d1 represents today's forecast that was generated yesterday, d2 represents today's forecast that was generated 2 days ago etc.

To evaluate and compare the performance of the models, the following statistical measures were used: mean absolute error (MAE), mean absolute percentage error (MAPE), mean square error (MSE), root-mean-square error (RMSE), and coefficient of determination ($R^2$) [18]. For reference, these metrics are defined as follows, given actual values $y_t$ and estimated values $\hat{y}_t$, at instants $t = 1, 2, \ldots, n$, and the mean value of the actual samples $\bar{y}$: MAE $= \frac{1}{n}\sum_{t=1}^{n}|y_t - \hat{y}_t|$, MAPE $= \frac{1}{n}\sum_{t=1}^{n}\frac{|y_t - \hat{y}_t|}{|y_t|} \times 100\%$, MSE $= \frac{1}{n}\sum_{t=1}^{n}(y_t - \hat{y}_t)^2$, RMSE $= \sqrt{MSE}$ and $R^2 = 1 - \sum_{t=1}^{n}(y_t - \hat{y}_t)^2/\sum_{t=1}^{n}(y_t - \bar{y})^2$. In this group of evaluation measures, MAE measures the average of the absolute differences between the original and predicted values, in the original unit, while MAPE measures the average absolute percentage error. MSE represents the average of the squared differences between the original and predicted values, which is a measure of the variance of the residuals in the squared unit of the original data. RMSE is the square root of MSE, which returns the residual variance to the original unit. Finally, $R^2$ stands for the proportion of the variance in the dependent variable that is explained by the linear regression model, and values should be close to 1 for a good model. On the other hand, values of MAE, MAPE, MSE, and RMSE should be close to 0 for a good model.

Depending on the type of problem being studied (estimating $ET_0$ or SR), two targets were considered: (i) (daily) $ET_0$ was calculated using the FAO56PM formula and input data from the WS, including temperature, humidity, wind speed, and solar radiation; (ii) the (daily) average solar radiation measured by the weather station was used as the target for solar radiation.

### 3.2 Exploratory data analysis of the forecast data

This section compares the historical weather forecast data, collected from Visual Crossing (VC) and OpenWeatherMap (OWM), from d0 up to d15, against the actual values measured by the Vale do Lobo WS, by means of the $R^2$ metric. Fig. **2** (see also **Table** 1 in the appendix) presents the evolution of the $R^2$ metric from d0 up to d15, for maximum and minimum temperature [TempMax, TempMin], average humidity [HumidityAvg], and average wind [WindAvg]. These are relevant weather parameters since they are used as model input features to the models used in this study.

Independent of the feature, as predictable since a weather forecast that is issued today should be more accurate than one that was generated 15 days ago, $R^2$ has its higher values for d0 and gradually degrades as it approaches d15. Further, the best $R^2$ values are obtained for maximum and minimum temperature, ranging between 0.772 and 0.346, and between 0.855 and 0.408, respectively. It is also observable that, for these features, the initial $R^2$ values start to drop more sharply after d3. Furthermore, the $R^2$ curves have similar behavior for minimum and maximum temperature, dropping below 0.7 after d7 and d5, respectively. However, from all the forecast weather parameters, these are the more reliable for midterm forecast. For example, the $R^2$ values for average wind speed and average humidity are inferior to 0 after d6 and d8, respectively. With respect to average humidity, $R^2$ is very close for both weather service providers, but VC has a slightly higher $R^2$ value up to d3, where both curves match.

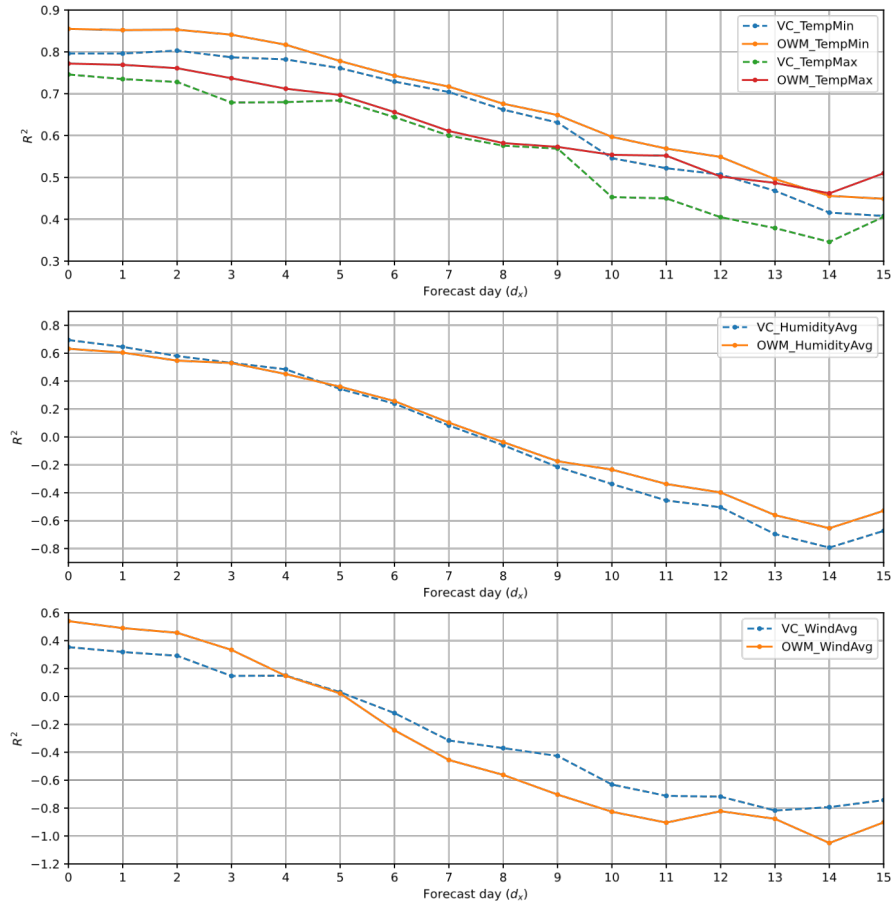

**Fig. 2.** Plots of the $R^2$ evolution for VC and OWM weather data in function of forecast day for maximum [TempMax] and minimum [TempMin] temperature (top), average humidity [HumidityAvg] (middle), and wind speed average [WindAvg] (bottom).

At d5, the $R^2$ value falls below 0.4, and at d8 falls below zero, which indicates that the average humidity forecasts should probably be used, at most, up to d3 or d4, for both services. Regarding average wind, starting at d0 with an $R^2$ of 0.540 and 0.354, for OWM and VC, respectively, both curves match at d4 with an $R^2$ below 0.2, indicating that from all the forecast weather parameters, average wind speed is the most unreliable, and should possibly be considered only up to d2. In general, for the Vale do Lobo WS coordinates, OWM has slightly higher $R^2$ values than VC (of course, this might have a different outcome for different locations). The impact of forecast values will be discussed further in the following sections.

### 3.3 Experimental setup

This study was performed in a computational environment using Python v3.9.7, Numpy v1.21.4 [19], Pandas v1.5.2 [20, 21], Tensor Flow v2.6.0 [22], Keras 2.6.0 [23], Scikit-learn v1.0.1 [24], and PyET v1.1.0 [25]. The Pandas library was employed for data analysis and manipulation, PyET to calculate $ET_0$ using the FAO56PM method, Scikit-learn for data preprocessing and model metrics evaluation. All neural network models presented were created using Keras, which operates on top of Tensor Flow. Computation was performed on a 2020 MacBook Air equipped with an Apple M1 SoC chip and 16 GB of RAM, running MacOS Big Sur v11.6.4.

## 4 Impact of using forecast data on $ET_0$ and SR Regressors

In this section the impact of employing weather forecast data as feature inputs for $ET_0$ and SR model inference is presented. To be clear, the models were not retrained using data from the WFSs, only inference was done. The models are thoroughly explained in [7], and were introduced in Section 1. More specifically, the following models for $ET_0$ inference are used: (i) a model that directly estimates $ET_0$ using an ANN model ($ET_0^{ANN}$), and (ii) a hybrid model that estimates SR using an ANN model ($SR^{ANN}$), and then uses that estimation for $ET_0$ computation, using the FAO56-PM method ($ET_0^{Hyb.}$). Table 2 (in appendix) presents the $R^2$, RMSE, MSE, MAE, and MAPE metrics for the estimation of $ET_0^{ANN}$, $SR^{ANN}$, and $ET_0^{Hyb.}$. Those values were obtained using as input to the models' historical weather forecast data from VC and OWM (for the Vale do Lobo WS coordinates), from d0 up to d15, as previously presented in Section 3.

Figure 3 shows the evolution (from d0 to d15) of the $R^2$ and MAPE plots for the $ET_0^{ANN}$ and $ET_0^{Hyb.}$, obtained by doing $ET_0$ model inference, while using meteorological forecast data from VC and OWM as model input features. In general, when looking at the $R^2$ plot (see Fig. 3, top), it can be concluded that up to d5, for both $ET_0^{ANN}$ and $ET_0^{Hyb.}$, $R^2$ is slightly higher when using OWM forecast data. Then, starting at d5 and up to d9 the curves match and have similar $R^2$ values. $R^2$ starts at d0 with values around 0.86 and 0.88 for VC and OWM, respectively, and falls below 0.8 after d6, indicating that for 6-day $ET_0$ estimation, somehow trustful results can be obtained using forecast data from both providers. It is also noted that the $ET_0^{Hyb.}$ model always gives better results than the $ET_0^{ANN}$ model. The $R^2$ for the $ET_0^{Hyb.}$ only falls below 0.7 after d11 and d13, for VC and OWM, respectively, suggesting that the proposed models can be employed for "mid" term $ET_0$ estimation, which can play a major role on the design of predictive irrigation scheduling algorithms. The MAPE plot (Fig. 3, bottom), shows that the mean absolute percentage error for each of the $ET_0$ models have similar values, when using either VC or OWM forecast data, with OWM only having a marginal advantage over VC. The MAPE plot also reinforces that the $ET_0^{Hyb.}$ model performs better than the $ET_0^{ANN}$ model, with a difference between both of at least 2.5 %.

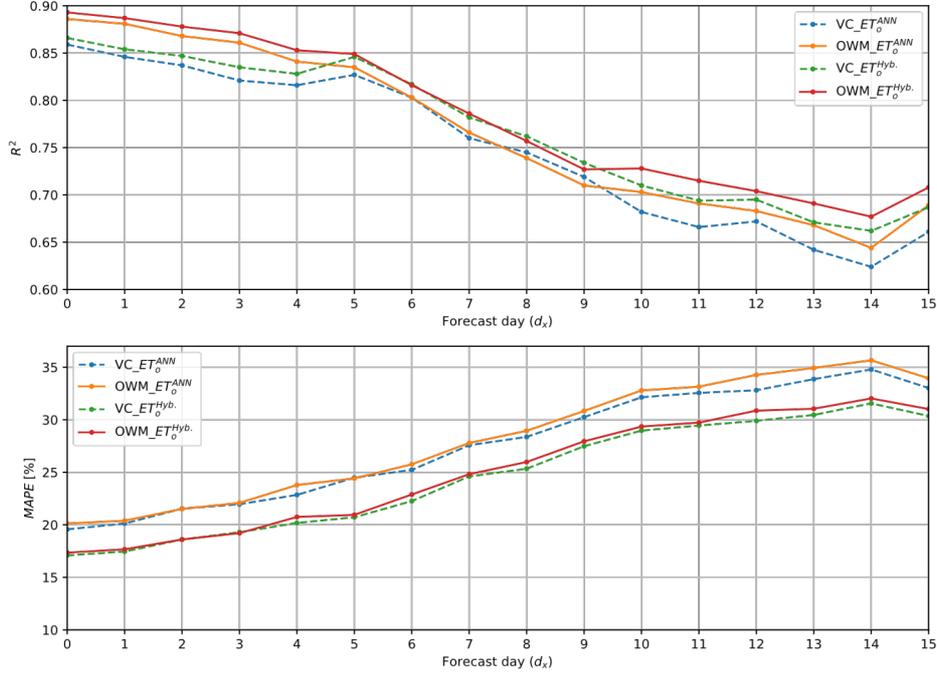

Fig.3: $R^2$ (top), and MAPE (bottom) plots for $ET_0^{ANN}$, and $ET_0^{Hyb.}$, using historical weather forecast data collected from VC and OWM, as model input features.

Considering 25 % as an admissible threshold for the MAPE, as before, a somehow trustful 6-day $ET_0$ estimation can be obtained using forecast data from both providers. Lowering the threshold to 20 % would suggest that the $ET_0^{Hyb.}$ predictions can be used for up to d4.

A set of violin plots of the MAE are presented on Fig. 4, where VC (blue), and OWM (orange) can be directly compared, for the $ET_0^{ANN}$ and $ET_0^{Hyb.}$, as well as the $SR^{ANN}$ models. When analyzing the plots, in general, and as expected, MAE increases as dx increases. It can also be seen that the models perform slightly better when using OWM data, but the difference is marginal.

These values are obviously worse than the ones reported by Vaz et al. [7] when using the observed (instead of forecast) Vale do Lobo WS test data. In that case, it was possible to reach an $R^2$ of 0.977 and 0.984, and a MAPE of 8.19 and 4.84 %, for the $ET_0^{ANN}$ and $ET_0^{Hyb.}$ models, respectively. To understand the impact of using weather forecast data on both $ET_0$ estimation models, it is important to compare these values with the ones obtained with d0 weather forecast data. The averaged metrics (between VC, and OWM), give an $R^2$ value of 0.873 and 0.880, and a MAPE of 19.85 and 17.23 %, for $ET_0^{ANN}$ and $ET_0^{Hyb.}$, respectively. The conclusion is that, for the Vale do Lobo WS, a rule of thumb for the impact of using weather forecast data, is an approximate reduction of 0.1 on $R^2$, and an increase of 12 % on the MAPE metric. Nevertheless, these values can improve the water efficiency of the green spaces by reducing or increasing the irrigation time according to the $ET_0$ predictions. Of course, this should also be integrated with precipitation, see Tab. 1 for an idea of the $R^2$ values associated to that forecast, being this out of the scope of this paper.

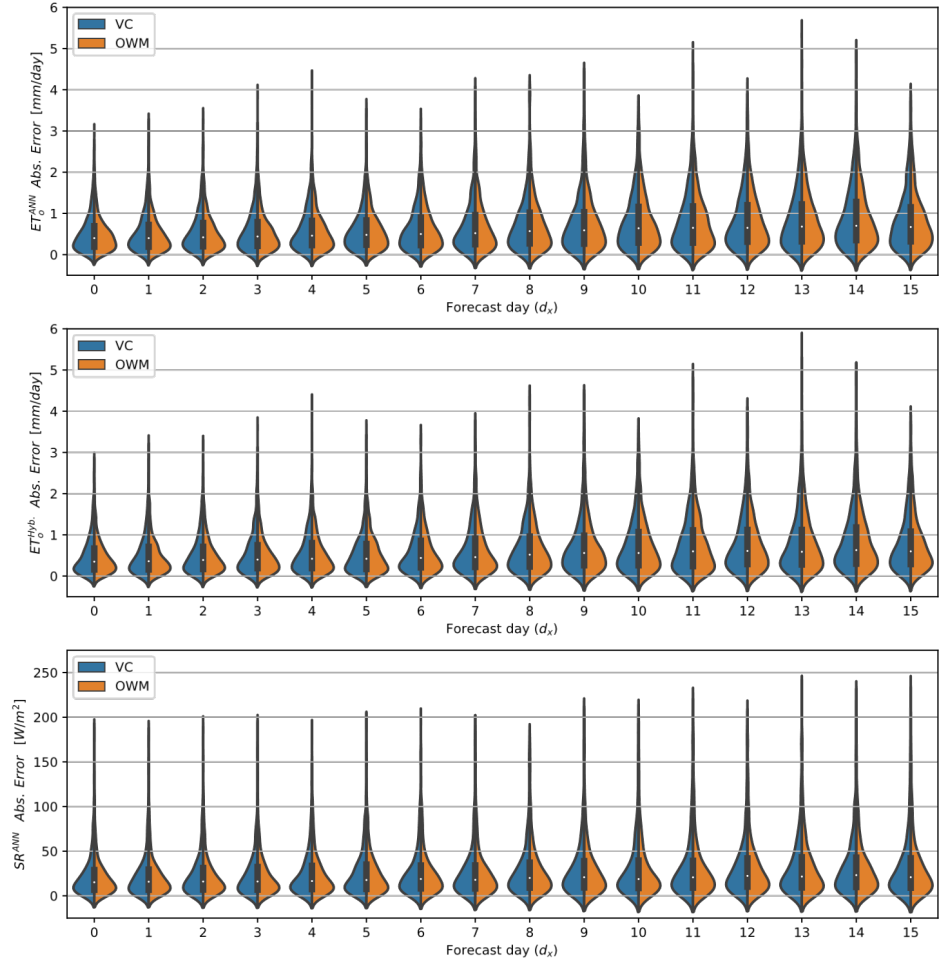

**Fig. 3.** Violin plots showing absolute error for $ET_0^{ANN}$ (top), $ET_0^{Hyb.}$ (middle), and $SR^{ANN}$ (bottom), in function of forecast day (d0, d1, ..., d15), and forecast data collected from VC and OWM.

## 5 Conclusion and future work

Evapotranspiration is a key parameter for the design of predictive irrigation algorithms, that take into account future crop water needs, as well as precipitation and other meteorological parameter forecasts. In this study, and to the best knowledge of the authors, the impact and comparison of using two different online WFSs that provide 15-day forecast data, was for the first time assessed, when applied as input features to $ET_0$ and SR estimation regressors based on deep learning. Despite the average performance of the metrics that are obtained when comparing the forecast meteorological data, with the data measured at the Vale do Lobo WS, especially regarding average humidity and wind, it was found that the $ET_0$ regressors still gave good performance metrics, and it was shown that they can be used for midterm $ET_0$ prediction, by using as input features data that is provided by common online and free WFSs. It was also concluded, that the $ET_0$ estimation models had similar performance, when employing either VC or OWM weather forecast data and that the hybrid $ET_0^{Hyb.}$ model has better performance than the $ET_0^{ANN}$ model, when using weather forecast data. This was expected and in line with previous findings by the authors.

Future work will include the integration of the precipitation forecast in a formula to compute the irrigation time, the generalization of these results to more WSs, as well as retrain the models using (d0) weather forecast data as features, instead of the WS measured features, to further try to improve $ET_0$ estimation model performance when using freely available online weather forecast services.

**Acknowledgements:** The authors thank the GSSIC Project's Companies, Visualforma - Tecnologias de Informação, S. A. and Itelmatis, Lda., and the Portuguese Foundation for Science and Technology (FCT) under Project UIDB/50009/2020—LARSyS, Project UIDB/00631/2020–CEOT BASE, Project UIDP/00631/2020—CEOT Programático, and Project UIDB/05183/2020–MED. We also wish to thank Visual Crossing for offering the data for this study.

## A    Summary of performance measures

**Table** 1. $R^2$ metric comparing historical weather forecast data collected from VC and OWM weather forecast services against the measured values at the Vale do Lobo WS.

|  | TempMax | | TempIn | | HumidityAvg | | WindAvg | | Precipitation | |
|---|---|---|---|---|---|---|---|---|---|---|
|  | VC | OWM | VC | OWM | VC | OWM | VC | OWM | VC | OWM |
| $d_0$ | 0.746 | 0.772 | 0.796 | 0.855 | 0.695 | 0.632 | 0.354 | 0.54 | -0.093 | -0.039 |
| $d_1$ | 0.735 | 0.769 | 0.796 | 0.852 | 0.646 | 0.605 | 0.319 | 0.49 | -0.438 | -0.111 |
| $d_2$ | 0.728 | 0.761 | 0.803 | 0.853 | 0.58 | 0.547 | 0.292 | 0.457 | -0.685 | -0.545 |
| $d_3$ | 0.679 | 0.737 | 0.787 | 0.841 | 0.531 | 0.53 | 0.147 | 0.334 | -0.614 | -0.407 |
| $d_4$ | 0.68 | 0.712 | 0.782 | 0.817 | 0.485 | 0.451 | 0.15 | 0.148 | -1.454 | -0.907 |
| $d_5$ | 0.684 | 0.697 | 0.761 | 0.778 | 0.344 | 0.36 | 0.031 | 0.022 | -0.451 | -0.367 |
| $d_6$ | 0.644 | 0.656 | 0.729 | 0.743 | 0.239 | 0.257 | -0.119 | -0.242 | -1.238 | -1.106 |
| $d_7$ | 0.600 | 0.611 | 0.704 | 0.717 | 0.082 | 0.103 | -0.315 | -0.455 | -1.57 | -1.388 |
| $d_8$ | 0.576 | 0.582 | 0.662 | 0.676 | -0.059 | -0.037 | -0.37 | -0.562 | -0.943 | -0.856 |
| $d_9$ | 0.569 | 0.573 | 0.631 | 0.649 | -0.215 | -0.174 | -0.426 | -0.703 | -1.145 | -1.006 |
| $d_{10}$ | 0.453 | 0.554 | 0.546 | 0.597 | -0.337 | -0.234 | -0.631 | -0.826 | -0.868 | -1.358 |
| $d_{11}$ | 0.45 | 0.552 | 0.522 | 0.569 | -0.455 | -0.337 | -0.712 | -0.904 | -0.962 | -1.299 |
| $d_{12}$ | 0.405 | 0.502 | 0.507 | 0.549 | -0.504 | -0.398 | -0.718 | -0.822 | -1.268 | -0.865 |
| $d_{13}$ | 0.379 | 0.487 | 0.468 | 0.496 | -0.696 | -0.56 | -0.817 | -0.877 | -1.898 | -2.273 |
| $d_{14}$ | 0.346 | 0.462 | 0.416 | 0.456 | -0.793 | -0.654 | -0.793 | -1.05 | -1.031 | -1.45 |
| $d_{15}$ | 0.406 | 0.51 | 0.408 | 0.449 | -0.673 | -0.529 | -0.743 | -0.902 | -1.283 | -1.306 |

Table 2. $R^2$, RMSE, MSE, MAE and MAPE metrics for $ET_0^{ANN}$, and $ET_0^{Hyb.}$, using historical weather forecast data collected from VC and OWM, as model input features. [Units for RMSE, MSE, and MAE are, respectively: *mm/day*, *mm²/day*, *mm/day* for $ET_o$, and $W/m^2/day$, $(W/m^2)^2/day$, $W/m^2/day$ for SR.]

| Forecast Day | Metrics | $ET_o^{ANN}$ VC | $ET_o^{ANN}$ OWM | $ET_o^{Hyb.}$ VC | $ET_o^{Hyb.}$ OWM | $SR^{ANN}$ VC | $SR^{ANN}$ OWM |
|---|---|---|---|---|---|---|---|
| $d_0$ | $R^2$ | 0.859 | 0.886 | 0.866 | 0.893 | 0.849 | 0.858 |
| | RMSE | 0.704 | 0.633 | 0.688 | 0.614 | 33.331 | 32.251 |
| | MSE | 0.496 | 0.401 | 0.473 | 0.378 | 1110.979 | 1040.145 |
| | MAE | 0.53 | 0.49 | 0.5 | 0.46 | 22.92 | 21.97 |
| | MAPE | 19.57 | 20.13 | 17.1 | 17.35 | 20.35 | 20.34 |
| $d_1$ | $R^2$ | 0.846 | 0.881 | 0.854 | 0.887 | 0.847 | 0.858 |
| | RMSE | 0.736 | 0.648 | 0.718 | 0.632 | 33.435 | 32.238 |
| | MSE | 0.542 | 0.42 | 0.515 | 0.399 | 1117.905 | 1039.272 |
| | MAE | 0.55 | 0.49 | 0.52 | 0.47 | 22.97 | 22.18 |
| | MAPE | 20.13 | 20.39 | 17.46 | 17.67 | 20.09 | 20.45 |
| $d_2$ | $R^2$ | 0.837 | 0.868 | 0.847 | 0.878 | 0.836 | 0.843 |
| | RMSE | 0.76 | 0.683 | 0.735 | 0.655 | 34.588 | 33.924 |
| | MSE | 0.577 | 0.466 | 0.54 | 0.429 | 1196.327 | 1150.869 |
| | MAE | 0.58 | 0.52 | 0.54 | 0.49 | 23.87 | 23.26 |
| | MAPE | 21.54 | 21.52 | 18.6 | 18.61 | 21.44 | 21.48 |
| $d_3$ | $R^2$ | 0.821 | 0.861 | 0.835 | 0.871 | 0.831 | 0.842 |
| | RMSE | 0.796 | 0.701 | 0.765 | 0.675 | 35.019 | 34.012 |
| | MSE | 0.634 | 0.492 | 0.585 | 0.455 | 1226.363 | 1156.814 |
| | MAE | 0.61 | 0.54 | 0.57 | 0.5 | 24.2 | 23.38 |
| | MAPE | 21.96 | 22.09 | 19.3 | 19.21 | 21.58 | 21.59 |
| $d_4$ | $R^2$ | 0.816 | 0.841 | 0.828 | 0.853 | 0.822 | 0.828 |
| | RMSE | 0.803 | 0.748 | 0.778 | 0.72 | 35.989 | 35.453 |
| | MSE | 0.645 | 0.56 | 0.605 | 0.519 | 1295.188 | 1256.935 |
| | MAE | 0.61 | 0.57 | 0.58 | 0.53 | 25.08 | 24.69 |
| | MAPE | 22.85 | 23.79 | 20.18 | 20.75 | 22.2 | 22.72 |
| $d_5$ | $R^2$ | 0.827 | 0.835 | 0.846 | 0.849 | 0.817 | 0.823 |
| | RMSE | 0.779 | 0.764 | 0.735 | 0.731 | 36.527 | 35.983 |
| | MSE | 0.607 | 0.584 | 0.541 | 0.534 | 1334.215 | 1294.809 |
| | MAE | 0.60 | 0.59 | 0.55 | 0.55 | 26.06 | 25.49 |
| | MAPE | 24.49 | 24.42 | 20.72 | 20.95 | 23.27 | 23.02 |
| $d_6$ | $R^2$ | 0.803 | 0.803 | 0.817 | 0.816 | 0.801 | 0.804 |
| | RMSE | 0.833 | 0.834 | 0.803 | 0.806 | 38.147 | 37.867 |
| | MSE | 0.694 | 0.695 | 0.645 | 0.65 | 1455.211 | 1433.914 |
| | MAE | 0.64 | 0.64 | 0.6 | 0.6 | 26.65 | 26.47 |
| | MAPE | 25.23 | 25.76 | 22.26 | 22.89 | 24.47 | 24.27 |
| $d_7$ | $R^2$ | 0.76 | 0.766 | 0.782 | 0.786 | 0.775 | 0.788 |
| | RMSE | 0.919 | 0.909 | 0.874 | 0.869 | 40.407 | 39.361 |
| | MSE | 0.845 | 0.826 | 0.765 | 0.756 | 1632.731 | 1549.282 |
| | MAE | 0.69 | 0.68 | 0.65 | 0.64 | 27.5 | 26.76 |
| | MAPE | 27.58 | 27.79 | 24.61 | 24.82 | 25.49 | 24.98 |
| $d_8$ | $R^2$ | 0.745 | 0.739 | 0.762 | 0.757 | 0.762 | 0.767 |
| | RMSE | 0.946 | 0.959 | 0.915 | 0.927 | 41.615 | 41.318 |
| | MSE | 0.896 | 0.921 | 0.838 | 0.859 | 1731.844 | 1707.188 |
| | MAE | 0.72 | 0.73 | 0.68 | 0.69 | 29.12 | 28.86 |
| | MAPE | 28.37 | 28.95 | 25.34 | 25.98 | 25.82 | 26.25 |
| $d_9$ | $R^2$ | 0.719 | 0.71 | 0.734 | 0.727 | 0.732 | 0.748 |
| | RMSE | 0.994 | 1.012 | 0.966 | 0.981 | 44.237 | 42.968 |

**Table 2.** (*cont.*) $R^2$, RMSE, MSE, MAE and MAPE metrics for $ET_0^{ANN}$, and $ET_0^{Hyb.}$, using historical weather forecast data collected from VC and OWM, as model input features. [Units for RMSE, MSE, and MAE are, respectively: *mm/day*, *mm²/day*, *mm/day* for $ET_o$, and *W/m²/day*, *(W/m²)²/day*, *W/m²/day* for SR.]

| Forecast Day | Metrics | $ET_o^{ANN}$ VC | $ET_o^{ANN}$ OWM | $ET_o^{Hyb.}$ VC | $ET_o^{Hyb.}$ OWM | $SR^{ANN}$ VC | $SR^{ANN}$ OWM |
|---|---|---|---|---|---|---|---|
| | MSE | 0.9888 | 1.025 | 0.9333 | 0.963 | 1956.933 | 1846.234 |
| | MAE | 0.75 | 0.76 | 0.72 | 0.73 | 30.45 | 29.61 |
| | MAPE | 30.25 | 30.84 | 27.48 | 27.95 | 28.98 | 27.24 |
| | $R^2$ | 0.682 | 0.703 | 0.71 | 0.728 | 0.709 | 0.727 |
| | RMSE | 1.055 | 1.024 | 1.006 | 0.98 | 45.961 | 44.679 |
| $d_{10}$ | MSE | 1.113 | 1.049 | 1.013 | 0.96 | 2112.418 | 1996.214 |
| | MAE | 0.81 | 0.79 | 0.76 | 0.74 | 31.2 | 30.1 |
| | MAPE | 32.14 | 32.79 | 28.97 | 29.35 | 29.19 | 28.53 |
| | $R^2$ | 0.666 | 0.691 | 0.694 | 0.715 | 0.702 | 0.714 |
| | RMSE | 1.084 | 1.044 | 1.038 | 1.004 | 46.551 | 45.76 |
| $d_{11}$ | MSE | 1.175 | 1.091 | 1.078 | 1.008 | 2166.994 | 2093.95 |
| | MAE | 0.83 | 0.81 | 0.78 | 0.76 | 31.76 | 30.96 |
| | MAPE | 32.56 | 33.15 | 29.45 | 29.73 | 28.92 | 28.55 |
| | $R^2$ | 0.672 | 0.683 | 0.695 | 0.704 | 0.718 | 0.725 |
| | RMSE | 1.075 | 1.058 | 1.037 | 1.023 | 45.367 | 44.865 |
| $d_{12}$ | MSE | 1.155 | 1.12 | 1.074 | 1.046 | 2058.132 | 2012.909 |
| | MAE | 0.83 | 0.83 | 0.79 | 0.78 | 32.04 | 31.45 |
| | MAPE | 32.81 | 34.27 | 29.9 | 30.87 | 28.99 | 28.67 |
| | $R^2$ | 0.642 | 0.668 | 0.671 | 0.691 | 0.703 | 0.719 |
| | RMSE | 1.122 | 1.082 | 1.075 | 1.044 | 46.483 | 45.346 |
| $d_{13}$ | MSE | 1.26 | 1.171 | 1.157 | 1.089 | 2160.701 | 2056.296 |
| | MAE | 0.86 | 0.84 | 0.8 | 0.79 | 32.66 | 31.61 |
| | MAPE | 33.87 | 34.93 | 30.46 | 31.05 | 29.84 | 29.36 |
| | $R^2$ | 0.624 | 0.644 | 0.662 | 0.677 | 0.7 | 0.713 |
| | RMSE | 1.15 | 1.121 | 1.09 | 1.068 | 46.587 | 45.818 |
| $d_{14}$ | MSE | 1.323 | 1.256 | 1.188 | 1.142 | 2170.357 | 2099.281 |
| | MAE | 0.9 | 0.87 | 0.84 | 0.82 | 33.14 | 32.25 |
| | MAPE | 34.79 | 35.66 | 31.56 | 32.03 | 29.87 | 29.73 |
| | $R^2$ | 0.661 | 0.689 | 0.687 | 0.708 | 0.696 | 0.704 |
| | RMSE | 1.091 | 1.048 | 1.049 | 1.015 | 46.978 | 46.57 |
| $d_{15}$ | MSE | 1.189 | 1.099 | 1.1 | 1.03 | 2206.964 | 2168.786 |
| | MAE | 0.84 | 0.82 | 0.8 | 0.78 | 32.18 | 31.63 |
| | MAPE | 33.02 | 33.95 | 30.36 | 31.02 | 29.97 | 29.86 |